\pdfoutput=1

\documentclass[11pt]{article}

\usepackage{acl}

\usepackage{times}
\usepackage{latexsym}

\usepackage[T1]{fontenc}

\usepackage[utf8]{inputenc}

\usepackage{microtype}

\usepackage{lipsum}

\usepackage{xspace}
\usepackage{booktabs}
\usepackage{graphicx}
\usepackage{enumitem}
\usepackage{xcolor}
\usepackage{tikz-dependency}
\usepackage{url}
\usepackage{amssymb}
\usepackage{float}
\usepackage{todonotes}
\usepackage{csquotes}
\usepackage{xspace}
\usepackage{listings}
\usepackage{caption}
\usepackage{subcaption}
\usepackage{multirow}
\usepackage{booktabs}

\setlist[description]{leftmargin=\parindent,labelindent=0pt}

\usepackage{tablefootnote}
\usepackage{tabularx}
\newcolumntype{L}{>{\raggedright\arraybackslash}X}
\usepackage{amsmath}
\usepackage{tikz}

\newcommand{\aref}[1]{Appendix~\ref{#1}}

\newcommand{\tref}[1]{Table~\ref{#1}}
\newcommand{\fref}[1]{Figure~\ref{#1}}

\newcommand{\corpusNameShort}{MuLMS-AZ\xspace}

\newcommand{\azLabelMotivation}{\textsc{Motivation}\xspace}
\newcommand{\azLabelBackground}{\textsc{Background}\xspace}
\newcommand{\azLabelPriorWork}{\textsc{PriorWork}\xspace}
\newcommand{\azLabelExperiment}{\textsc{Experiment}\xspace}
\newcommand{\azLabelPreparation}{\textsc{Preparation}\xspace}
\newcommand{\azLabelCharacterization}{\textsc{Characterization}\xspace}
\newcommand{\azLabelExplanation}{\textsc{Explanation}\xspace}

\newcommand{\azLabelResults}{\textsc{Results}\xspace}
\newcommand{\azLabelConclusion}{\textsc{Conclusion}\xspace}

\newcommand{\azLabelAbstract}{\textsc{Abstract}\xspace}
\newcommand{\azLabelCaption}{\textsc{Caption}\xspace}
\newcommand{\azLabelHeading}{\textsc{Heading}\xspace}
\newcommand{\azLabelMetadata}{\textsc{Metadata}\xspace}

\newcommand{\pubmedLabelObjective}{\textsc{Objective}\xspace}
\newcommand{\pubmedLabelBackground}{\textsc{Background}\xspace}
\newcommand{\pubmedLabelResults}{\textsc{Results}\xspace}
\newcommand{\pubmedLabelConclusions}{\textsc{Conclusions}\xspace}

\newcommand{\totalNumSentences}{10186\xspace} %

\makeatletter
\def\url@leostyle{%
  \@ifundefined{selectfont}{\def\UrlFont{\sf}}{\def\UrlFont{\scriptsize\sffamily}}}
\makeatother
\DeclareUnicodeCharacter{03B3}{$\gamma$}
\DeclareUnicodeCharacter{03C1}{$\rho$}

\title{\corpusNameShort: An Argumentative Zoning Dataset\\ for the Materials Science Domain}

\author{Timo Pierre Schrader$^{1,6}$~
  Teresa Bürkle$^2$~
  Sophie Henning$^{1,3}$~
  Sherry Tan$^4$~
  Matteo Finco$^2$\\
  {\bf Stefan Grünewald$^{1,5}$}~
  {\bf Maira Indrikova$^2$}~
  {\bf Felix Hildebrand$^2$}~
  {\bf Annemarie Friedrich$^6$} \\
  $^1$Bosch Center for Artificial Intelligence, Renningen, Germany \\ 
    $^2$Robert Bosch GmbH, Stuttgart, Germany \hspace{-1.2mm} $^3$LMU Munich, Germany \hspace{-1.2mm} $^4$TU Darmstadt, Germany \\ $^5$University of Stuttgart, Germany \hspace{2mm} $^6$University of Augsburg, Germany \\
\texttt{timo.schrader|teresa.buerkle|sophie.henning@de.bosch.com} \\
  \texttt{annemarie.friedrich@informatik.uni-augsburg.de}}

\date{}

\begin{document}
\maketitle
\begin{abstract}
Scientific publications follow conventionalized rhetorical structures.
Classifying the \textit{Argumentative Zone} (AZ), e.g., identifying whether a sentence states a \textsc{Motivation}, a \textsc{Result} or \textsc{Background} information, has been proposed to improve processing of scholarly documents.
In this work, we adapt and extend this idea to the domain of materials science research.
We present and release a new dataset of 50 manually annotated research articles.
The dataset spans seven sub-topics and is annotated with a materials-science focused multi-label annotation scheme for AZ.
We detail corpus statistics and demonstrate high inter-annotator agreement.
Our computational experiments show that using domain-specific pre-trained transformer-based text encoders is key to high classification performance.
We also find that AZ categories from existing datasets in other domains are transferable to varying degrees.
\end{abstract}

\section{Introduction}
\label{sec:intro}
In academic writing, it is custom to adhere to a rhetorical argumentation structure to convince readers of the relevance of the work to the field \citep{swales1990discourse}.
For example, authors typically first indicate a gap in prior work before stating the goal of their own research.
\textit{Argumentative Zoning} (AZ) is a natural language processing (NLP) task in which sentences are classified according to their argumentative roles with varying granularity \citep{teufel-etal-1999-annotation, teufel-etal-2009-towards}.
AZ information can then be used for summarization \citep{teufel-moens-2002-articles, el-ebshihy-etal-2020-artu}, improved citation indexing \citep{teufel2006argumentative}, or writing assistance \citep{feltrim2006argumentative}.

\begin{table}[t]
\vspace{3mm}

\footnotesize
    \centering
    \begin{tabular}{lr|lr}
    \toprule
    \textbf{Label} & \textbf{Count} & \textbf{Label} & \textbf{Count}\\
    \midrule
        \azLabelMotivation & 363 & \azLabelExplanation & 603 \\ 
        \azLabelBackground & 3155 & \azLabelResults & 2953 \\
        - \azLabelPriorWork & 1824 & \azLabelConclusion & 680 \\
\azLabelExperiment & 2579
& \azLabelHeading & 702\\
 - \textsc{Prep.} & 962
         & \textsc{Caption} & 485\\
         - \textsc{Charact.} & 1347 & \azLabelMetadata & 210 \\
        \bottomrule
\end{tabular}
    \caption{\corpusNameShort label counts (multi-label).}
    \label{tab:az_labelcounts}
\end{table}

Manually annotated AZ datasets \citep{teufel-etal-1999-annotation,fisas-etal-2016-multi,soldatova2007ontology} only exist for few domains and employ differing annotation schemes.
The resulting models are not directly applicable to our domain of interest, materials science research, which presents a challenging domain for current NLP methods \citep[e.g.,][]{mysore-etal-2019-materials,friedrich-etal-2020-sofc,ogorman-etal-2021-ms}.
In this paper, we present \corpusNameShort, the first dataset annotated for AZ in this domain.
Working together with domain experts, we derive a hierarchical multi-label \textbf{annotation scheme} (see \tref{tab:az_labelcounts}).
Our scheme includes domain-specific labels such as descriptions of the materials' \azLabelPreparation and \azLabelCharacterization, which are crucial distinctions also for NLP applications from the domain experts' view.

This \textbf{resource paper} makes the following contributions:

\begin{itemize}[leftmargin=*]
 \setlength\itemsep{0pt}
    \item We present a \textbf{dataset} of 50 scientific articles (more than 10,000 sentences) in the domain of materials science manually annotated by domain experts with a hierarchical fine-grained \textbf{annotation scheme} for AZ with high agreement. The corpus will be publicly released.\footnote{\url{https://github.com/boschresearch/mulms-az-codi2023}}
    \item We apply several neural models to our dataset that will serve as strong baselines for future work using our new benchmark. We find (a) that using domain-specific pre-trained transformers is key to a successful model, (b) that multi-task learning with existing AZ datasets leads to small benefits, and (c) that the effectiveness of transfer learning of materials science AZ labels from other corpora differs by label.
\end{itemize}

\section{Related Work}
\label{sec:relwork}
In this section, we describe related work on AZ.

\textbf{AZ Datasets.}
\tref{tab:dataStats} shows the statistics of several related datasets.
Three larger-scale datasets manually annotated with AZ information are the AZ-CL corpus \citep{teufel-etal-1999-annotation, teufel-moens-1999-discourse}, consisting of computational linguistics papers, the Dr. Inventor Multi-Layer Scientific Corpus \citep[DRI,][]{fisas-etal-2016-multi,fisas-etal-2015-discoursive}, featuring computer graphics papers, and, closest to our domain, the ART corpus \citep{soldatova2007ontology}, covering topics in physical chemistry and biochemistry.
\aref{sec:datasets-appendix} explains these datasets in more detail.
\citet{teufel-etal-2009-towards} also apply and adapt the annotation scheme of the AZ-CL corpus to the chemistry domain. %
\citet{accuosto2021argumentation} label sentences with argumentation-related categories (e.g., \textit{proposal}, \textit{means}, \textit{observation}).
Making use of sentence-wise author-provided keywords, a dataset of about 388k abstracts with silver standard rhetorical role annotations has been derived from PubMed/MEDLINE \citep{deMoura2018using}.

\textbf{Modeling.}
AZ has been modeled as a sentence classification task using maximum entropy models \citep{teufel2009robust}, SVMs, and CRFs \citep{guo-etal-2011-weakly} leveraging a variety of word, grammatical, heuristic, and discourse features \citep{guo-etal-2013-improved}.
Ensemble-based classifiers have also been shown to be effective \citep{badie2018zone,asadi2019automatic}.
LSTM-based models relying on word embeddings have been applied to AZ and to the fundamentally very similar task of assigning zones to sentences in job ads \citep{DBLP:journals/corr/Liu17a,deMoura2018using,gnehm-clematide-2020-text}.
BERT-style \citep{devlin-etal-2019-bert} models work well for AZ \citep{mo2020deep,brack2022cross}.
Multi-task training has been found to be beneficial for these models both in-domain \citep{lauscher-etal-2018-investigating} as well as cross-domain \citep{brack2021cross_domain}.

\textbf{Datasets in the Materials Science Domain.}
Several datasets targeting the domain of materials science research have recently been released.
\citet{mysore-etal-2019-materials} annotate paragraphs describing synthesis procedures with graph structures capturing relations and typed arguments.
\citet{friedrich-etal-2020-sofc} mark similar graph structures corresponding to experiment information in 45 open-access publications.
Several works and datasets address named entity recognition in the domain \citep{yamaguchi-etal-2020-sc,ogorman-etal-2021-ms}.

\begin{table}[t]
	\centering
	\small
	\setlength\tabcolsep{5pt}
	\begin{tabular}{lrrrr}
	\toprule
	& \textbf{AZ-CL} & \textbf{ART} & \textbf{DRI} & \textbf{\corpusNameShort}\\
	\midrule
		\# docs & 80 & 225 & 40 & 50\\
		\# sents & 12818 & 34995 & 10784 & \totalNumSentences\\
		\# labels & 7 & 11 & 10 & 12\\
	\bottomrule
	\end{tabular}
\caption{Manually annotated AZ corpora.}
\label{tab:dataStats}
\end{table}

\begin{table*}[htb]
    \centering
    \footnotesize
    \setlength\tabcolsep{4pt}
    \begin{tabular}{l|l|l}
    \toprule
       \textbf{Label}  & \textbf{Description} & \textbf{Example}\\
    \midrule
    \azLabelMotivation & aims/motivation of the study & \textit{In this study, we perform a systematic analysis of ...} \\
	\azLabelBackground & textbook-like technical background & \textit{The method is based on the Kelvin equation.}\\
	- \azLabelPriorWork & specific prior work relevant to current study & \textit{Irmawati et al. has concluded that ...}\\
	\azLabelExperiment & description of the experiment & \textit{We evaluate PtCo nanoparticle catalyst ...}\\
	 - \azLabelPreparation & steps describing the preparation of samples & \textit{The mixture was subjected to stirring for 60 minutes.}\\
	 - \textsc{Charact.} & characterizations and characterization & \textit{Ni foam surface coverage of the WO3 thin film and its} \\
	 & techniques of the involved materials & \textit{homogeneity were analyzed by energy–dispersive X-ray}\\
	 &  & \textit{spectroscopy (EDS).}\\
	\azLabelExplanation & statements (hypotheses or assumptions) & \textit{In our calculation, all Pt loadings were considered}\\
	& relevant to results or experimental settings & \hspace*{3mm}\textit{to be electrochemically active.}\\
	\azLabelResults & details on experimental results & \textit{The hydrogen adsorption/desorption peak is at about 0.2V.}\\
	\azLabelConclusion & conclusions and take-aways & \textit{This result indicated that ...}\\
    \bottomrule
    \end{tabular}
    \caption{Content-based \corpusNameShort Argumentative Zoning sentence labels.}
    \label{tab:az}
\end{table*}

\section{Data Sources and Annotated Corpus}
\label{sec:data}
\label{sec:dataset}
In this section, we present our new dataset.

\textbf{Source of Texts and Preprocessing.}
We select 50 scientific articles licensed under CC-BY from seven sub-areas of materials science: electrolysis, graphene, polymer electrolyte fuel cell (PEMFC), solid oxide fuel cell (SOFC), polymers, semiconductors, and steel.
The four SOFC papers were selected from the SOFC-Exp corpus \cite{friedrich-etal-2020-sofc}.
11 papers were selected from the OA-STM corpus\footnote{\url{https://github.com/elsevierlabs/OA-STM-Corpus}} and classified into the above subject areas by a domain expert.
The majority of the papers were found via PubMed\footnote{\url{https://pubmed.ncbi.nlm.nih.gov/}} and DOAJ\footnote{\url{https://doaj.org/}} using queries prepared by a domain expert.
For the OA-STM data, we use the sentence segmentation provided with the corpus, which has been created using GENIA tools \citep{tsuruoka-tsujii-2005-bidirectional}.
For the remaining texts, we rely on the sentence segmentation provided by INCEpTION v21.0 \citep{klie-etal-2018-inception} with some manual fixes.

\textbf{Annotation Scheme.}
AZs are functional sentence types, i.e., they capture the rhetorical function of a sentence. %
Together with several domain experts, we design a hierarchical scheme tailored to the materials science domain as shown in \tref{tab:az}.
In addition, we provide \azLabelAbstract, \azLabelHeading, \textsc{Metadata}, \textsc{Caption}, \textsc{Figure/Table} annotations for structural information.
We assume a multi-label setting in which annotators may assign any number of labels to a sentence.
Our detailed guidelines are available with our dataset. %

\textbf{Corpus Statistics.}
Documents are rather long (on average 203.7 sentences per document with a standard deviation of $\pm$73.2).
There is a tendency towards long sentences (28.7 tokens per sentence on average), but with high variation of $\pm$17.9 due to, e.g., short headings.
\tref{tab:az_labelcounts} shows how often each AZ label occurs. %
When ignoring tags for structural information %
8133 sentences have exactly one AZ label (or the AZ label and its supertype),
1056 sentences have two labels, and 11 sentences have 3 labels.
Labels are similarly distributed across data splits (see  \aref{sec:appendix-stats}).

\textbf{Inter-Annotator Agreement.}
Our entire dataset has been annotated by a single annotator, a graduate student of materials science, who was also involved in the design of the annotation scheme.
We compare the annotations of this main annotator to those of another annotator who holds a Master's degree in materials science and a PhD in engineering.
The agreement study is performed on 5 documents (357 sentences).
Due to the multi-label scenario, following \citet{krippendorff1980krippendorff} we measure $\kappa$ \citep{cohen1960kappa} for each label separately, comparing whether each annotator used a particular label on an instance or not (see \tref{tab:iaa_az}).
Our annotators achieve \enquote{substantial} agreement \citep{landis1977application} on most labels, \enquote{perfect} agreement on identifying \azLabelHeading{}s (see also \aref{sec:appendix-stats}).
Lower, though still \enquote{moderate}, agreement on \azLabelMotivation, \azLabelExplanation and \azLabelConclusion can in part be explained by their lower %
frequency
which makes it generally harder to obtain high $\kappa$-values.
Intuitively, they also have a more difficult nature compared to the other tags, e.g., we observe disagreements regarding what constitutes a \azLabelMotivation or an \azLabelExplanation versus what is purely reporting \azLabelBackground.
The full confusion matrix and a discussion of agreement on subtags are given in \aref{sec:appendix-stats}; a discussion of multi-label examples can be found in \aref{sec:examples}.

Our scores are in the same ballpark as those reported by \citet{teufel-etal-1999-annotation} %
on a similar annotation task.
For their 7-way task, they report $\kappa$ scores around 0.71-0.75, with differences between categories in one-vs-all measurements
ranging from about 0.49 to 0.78.
In sum, we conclude that agreement on AZ is satisfactory in our dataset.

\begin{table}[ht]
    \centering
    \footnotesize
    \begin{tabular}{lr|lr}
    \toprule
    \textbf{AZ Label} & $\kappa$ &  \textbf{AZ Label} & $\kappa$\\
    \midrule
        \azLabelHeading & 0.89 & \azLabelMetadata & 0.76 \\
        \azLabelMotivation & 0.44 & \azLabelBackground & 0.75 \\
        \azLabelConclusion & 0.55 & \azLabelExperiment & 0.78\\
        \azLabelExplanation & 0.39 & \azLabelResults & 0.70\\
       \bottomrule
    \end{tabular}
    \caption{IAA for AZ on 357 sentences.}   \label{tab:iaa_az}
\end{table}

 \section{Modeling}
\label{sec:models}
We model AZ as a multi-label classification problem, %
using BERT \citep{devlin-etal-2019-bert} as the underlying text encoder.
We also test domain-specific pre-trained variants of BERT.
SciBERT \citep{beltagy-etal-2019-scibert} has been pre-trained on articles in the scientific domain.
MatSciBERT \citep{gupta_matscibert_2022} is a version of SciBERT further pre-trained on materials science articles.
We use the CLS embedding as input to a linear layer, transform logits using a sigmoid function and choose labels if their respective score exceeds 0.5.
For multi-task experiments with other datasets, we use a single shared language model and one linear output layer per dataset.
For hyperparameters, see \aref{sec:hyperparameters}.

As shown in \tref{tab:az_labelcounts}, the dataset suffers from strong class imbalance.
Classifiers tend to underperform on minority labels \citep{johnson2019survey}.
To address this problem, we apply the \textbf{multi-label random oversampling}  \citep[ML-ROS,][]{charte2015mlros} algorithm during training.
The main idea behind ML-ROS is to dynamically duplicate instances of minority classes while taking the multi-label nature of the problem into account.
In a nutshell, the algorithm performs several oversampling iterations, keeping track of the imbalance ratios associated with each label and choosing instances that carry minority labels until a predefined number of additional samples have been chosen.
Details are given in \aref{sec:ml-ros}.

\section{Experimental Results}
\label{sec:experiments}
We here detail our experimental results.

\textbf{Settings.}
We split our corpus into train, dev, and test sets of 36, 7, and 7 documents. %
For all experiments and for hyperparameter tuning, we always train five models.
The training data is split into five folds.
Similar to cross-validation, we train on four folds and use the fifth fold for model selection \citep[cf.][]{van-der-goot-2021-need}, repeating this process five times (also for hyperparameter tuning).
The dev set is only used for tuning, and we report scores for the five models on test.
In this setting, deviations are naturally higher than when reporting results for the same training data.
For hyperparameters and implementation details, see \aref{sec:hyperparameters}.
To evaluate our experiments, we use hierarchical precision, recall, and F1 \citep{silla2011survey}.
These scores operate on the sets of labels per instance, always including the respective supertypes of gold or predicted labels.

\begin{table}[!t]
	\footnotesize
	\centering
	\setlength\tabcolsep{4pt}
	\begin{tabular}{llrrr}
		\toprule
		\textbf{Method} & \textbf{LM} & \textbf{mic.-F1} & \textbf{mac.-F1}\\ 
		\midrule
		No Oversampling & BERT & 72.6$_{\pm 1.0}$ & 65.5$_{\pm 0.7}$ \\
		& MatSciBERT & 76.3$_{\pm 0.7}$ & 70.1$_{\pm 0.7}$ \\
		& SciBERT & 76.2$_{\pm 0.9}$ & 70.2$_{\pm 0.6}$ \\
		\midrule
		ML-ROS & SciBERT & 76.7$_{\pm 0.7}$ & 70.6$_{\pm 0.9}$ \\
		+ MultiTask ART & SciBERT & 75.0$_{\pm 0.9}$ & 68.9$_{\pm 1.1}$\\
		+ MultiTask AZ-CL & SciBERT & \textbf{77.2$_{\pm 0.3}$} & \textbf{71.1$_{\pm 0.5}$} \\
		\midrule
		\textit{human agreement}* & & \textit{78.7}\hspace*{5.5mm} & \textit{74.9}\hspace*{5.5mm}\\
		\bottomrule
	\end{tabular}
	\caption{AZ classification results on \corpusNameShort test set. *Not directly comparable: computed on documents from agreement study (see \aref{sec:appendix-stats}).}
	\label{tab:az_results-mini}
\end{table}

\textbf{Results.}
\tref{tab:az_results-mini} shows the performance of our neural models on \corpusNameShort.
Overall, the categories can be learned well, approaching our estimate of human agreement.
SciBERT clearly outperforms BERT, i.e., using domain-specific embeddings is a clear advantage.
However, MatSciBERT does not add upon SciBERT.
We hence conduct the remaining experiments using SciBERT.
Using ML-ROS results in minor increases for most labels (see also \aref{sec:app-more-results}).
When multi-task learning with the AZ-CL dataset (using $40\%$ of its samples), further increases are observed.
It is worth noting that multi-task training with ART does not result in increases although the chemistry domain should be much closer to our domain.
This might indicate that despite the domain discrepancy, AZ annotations in AZ-CL are more compatible with ours.

\begin{table}[t]
    \centering
    \footnotesize
    \setlength\tabcolsep{5pt}
    \begin{tabular}{llll}
    \toprule
        \textbf{Training data} & \textbf{PM Label} & \hspace*{1mm}\textbf{P} & \hspace*{1mm}\textbf{R}\\
        \midrule
        PM, AZ-CL, ART, DRI & \textsc{Objective} & 36.1 & 28.3 \\
        PM, AZ-CL, ART, DRI & \textsc{Background} & 84.2 & 40.0 \\
        PM, ART, DRI & \textsc{Method} & 58.1 & 74.7  \\
        PM, ART, DRI & \textsc{Result} & 82.4 & 30.9\\
        PM, ART, DRI & \textsc{Conclusion} & 43.5 & 29.9 \\
        \hline
         \corpusNameShort & \textsc{Objective} & 56.8 & 54.3\\
        \corpusNameShort & \textsc{Background} & 82.1 & 78.8 \\
        \corpusNameShort & \textsc{Method} & 79.9 & 78.2 \\
        \corpusNameShort & \textsc{Result} & 82.1 & 83.2 \\
        \corpusNameShort & \textsc{Conclusion} & 43.5* &  29.9*\\
        \bottomrule
    \end{tabular}
    \caption{Results for transfer learning experiment. Precision and recall on \corpusNameShort test set. *not a typo.}
    \label{tab:transfer-experiment-mini}
\end{table}

As a first step to explaining what part of rhetorical information can be induced based only on data from other corpora, we perform a transfer learning experiment.
We carefully manually map the AZ labels of the various datasets (see \aref{sec:datasets-appendix}) to the coarse-grained categories used by PubMed (PM).
Using these mapped labels, we train binary classifiers that aim to detect the presence of a particular PM label.
As training data, we use ART, DRI, and a selection of documents from the PM dataset by \citet{deMoura2018using} that were published in materials science journals (see \aref{sec:journal-list}).
We add AZ-CL to the training data only if an unambiguous mapping of its categories to the PM label in question is possible.
Here, we use the dev set of \corpusNameShort for model selection and hyperparameter tuning.
Results for running the resulting classifiers on \corpusNameShort are reported in \tref{tab:transfer-experiment-mini}.
For \azLabelBackground and \azLabelResults, we observe a high precision, which indicates that similar rhetorical elements may be used.
\textsc{Objective} and \textsc{Method} seem to differ most across datasets as they are likely very domain- and problem-specific.
When training with mapped labels on the entire \corpusNameShort, we observe much higher recall scores across all label groups, again indicating the usefulness of our in-domain training data.

\section{Conclusion and Outlook}
\label{sec:conclusion}
We have presented a new AZ corpus in the field of materials science annotated by domain experts with high agreement.
Our experimental results demonstrate that strong classifiers can be learned on the data and that AZ labels can be transferred from related datasets only to a limited extent.

Our new dataset opens up new research opportunities on cross-domain AZ, class imbalance scenarios, and integrating AZ information in information extraction tasks in materials science.

\section*{Limitations}
This resource paper describes the dataset in detail, providing strong baselines and first initial cross-domain experiments.
It does not aim to provide an extensive set of experiments on cross-domain argumentative zoning yet.

The entire dataset is only singly-annotated.
The agreement study was performed on complete documents and hence has only limited data for several labels.
Due to the limited funding of the project, we could double-annotate the entire dataset.

Finally, we only test one model class (BERT-based transformers).
A potential next step is to test a bigger variety of models and embeddings.
Because AZ labels are interdependent within a document, especially document-level models or CRF-based models are promising methods to try.
We have also tested only one method (multi-label random oversampling) to deal with the strong class imbalance in the dataset.
We have not yet tested further such methods \citep{henning-etal-2023-survey} or data augmentation methods.

\section*{Ethical Considerations}
We took care of potential license issue of the data underlying our dataset by exclusively selecting open-access articles published under CC BY.

The main annotator was paid above the minimum wage of our country in the context of a full-time internship.
The annotator was aware of the goal of the study and consents to the public release of the data.
The remaining domain experts participated on a voluntary basis due to their interest in the topic.

\bibliography{custom,anthology}

\clearpage
\section*{Appendix}
\appendix
\section{Hyperparameters}
\label{sec:hyperparameters}

We implement all our models using PyTorch.
We use AdamW \citep{loshchilov2019decoupled} as the optimizer for all our models and set the batch size to $16/32$ depending on what works best and GPU restrictions.
The learning rate stays constant after a linear warmup phase.
We set a dropout rate to $0.1$ for the linear layer that takes the contextualized embeddings that are produced by BERT as input.
Early stopping is applied if the micro-F1 score has not improved for more than 3 epochs.
Binary cross entropy is the loss function for the \corpusNameShort output layer, whereas cross entropy is the loss function used for optimizing the multi-task output heads corresponding to the other AZ datasets.
\tref{tab:learning_rates} lists the various learning rates found during grid search.
We tested different learning rates between 1e-4 and 1e-7.
A refinement of the grid was done after an initial search, which almost always leads to a second search area within the range of 1e-6 to 9e-6.
When using ML-ROS, we oversample by $20\%$.
Training was performed on a single Nvidia A100 GPU or alternatively V100 GPU.

\begin{table}[ht]
	\footnotesize
	\centering
	\setlength\tabcolsep{2pt}
	\begin{tabular}{llr}
		\toprule
		\textbf{Method} & \textbf{LM} & \textbf{Learning Rate}\\ 
		\midrule
		No Oversampling & BERT & 3e-6 \\
		& MatSciBERT & 8e-6\\
		& SciBERT & 3e-6 \\
		\midrule
		ML-ROS & SciBERT & 2e-6 \\
		+ MT (+PM) & SciBERT & 7e-6 \\
		+ MT (+ART) & SciBERT & 2e-6 \\
		+ MT (+AZ-CL) & SciBERT & 2e-6 \\
		+ MT (+DRI) & SciBERT & 1e-6 \\
		+ MT (+ART+AZ+DRI) & SciBERT & 8e-6 \\
		Data Augm. (+PM) &  SciBERT & 8e-6 \\
		\bottomrule
	\end{tabular}
	\caption{Learning rates of the different model reported in \tref{tab:az_results}}
	\label{tab:learning_rates}
\end{table}

\section{Multi-Label Random Oversampling (ML-ROS) Algorithm}
\label{sec:ml-ros}
\fref{fig:pseudocode-ml-ros} details our adaption of the multi-label random oversampling (ML-ROS) algorithm originally proposed by \citet{charte2015mlros}.
In the initialization (lines 3-7), for each label, all the instances that carry a particular label are collected in what \citeauthor{charte2015mlros} call a \textit{bag}.
The main part of the algorithm (lines 10-24) does the following:
For each label $y$, the \textit{Imbalance Ratio per label} ($\mathrm{IRLbl}$), which is the ratio between the count of the most frequent label and the count of $y$, is calculated:
\begin{gather*}
    \mathrm{IRLbl}(y) = \frac{\max_{y'\in L} \sum_{i=1}^{|D|} h(y',Y_i)}{\sum_{i=1}^{|D|} h(y,Y_i)}
\end{gather*}
$D$ is the dataset, $L$ is the label set, $Y_i$ is the set of labels assigned to the $i$-th sample and $h$ is an indicator function evaluating if $y \in Y_i$.
Hence, the larger the value, the less frequently $y$ occurs compared to the most frequent label.

The per-label values are then used to determine the \textit{mean imbalance ratio} ($\mathrm{MeanIR}$):
\begin{gather*}
    \mathrm{MeanIR} = \frac{1}{|L|} \sum_{y'\in L} \mathrm{IRLbl}(y')
\end{gather*}

For each of the labels with an imbalance ratio exceeding the current MeanIR, a random instance of this label is duplicated.

The main part is repeated until the pre-specified size of the oversampled dataset is reached.
Our implementation differs from \citeauthor{charte2015mlros} in that we update $\mathrm{meanIR}$ in each iteration step and also oversample labels originally not being a minority label when their $\mathrm{IRLbl}$ exceeds $\mathrm{MeanIR}$ at the beginning of an iteration step.

\lstset{language=Python,mathescape=true, numbers=left, numberstyle=\tiny, morecomment=[s][\small]{/*}{*/}}

\begin{figure*}
\footnotesize
\centering
\begin{minipage}{0.7\textwidth}
\begin{lstlisting}
Inputs: <Dataset> D, <Percentage> P
Outputs: Oversampled dataset
$samplesToDuplicate$ <-- $|D|/100*P$ # P %
$L$ <-- labelsInDataset($D$) # Obtain the full set of labels
for each $label$ in $L$ do # Bags of samples for each label
    $Bag_{label}$ <-- getSamplesPerLabel($label$)
end for

while $samplesToDuplicate$ > 0 do # Loop duplicating instances
    $MeanIR$ <-- calculateMeanIR($D, L$)
    # Gather minority bags (bag: all instances of a given label)
    $minBags$ = []
    for each $label$ in $L$ do
        $IRLbl_{label}$ <-- calculateIRperLabel($D, label$)
        if $IRLbl_{label}$ > $MeanIR$ then
            $minBags$ += $Bag_{label}$
        end if
    end for
    # Duplicate a random sample from each minority bag
    for each $minBag_i$ in $minBags$ do
        $x$ <-- random($1, |minBag_i|$)
        duplicateSample($minBag_i, x$)
        $--samplesToDuplicate$
    end for
end while
\end{lstlisting}
\end{minipage}
\caption{Pseudocode for adapted (dynamic) ML-ROS algorithm.}
\label{fig:pseudocode-ml-ros}
\end{figure*}

\section{List of Materials Science Journals}
\label{sec:journal-list}
We used the list of materials-science related journals collected on Wikipedia to filter for abstracts in the PubMed Medline corpus published in journals.\footnote{ \url{https://en.wikipedia.org/w/index.php?title=List_of_materials_science_journals&oldid=1078212543}}

\section{Further Corpus Statistics for \corpusNameShort}
\label{sec:appendix-stats}
\tref{tab:AZ_label_counts} gives the counts of sentences carrying a particular AZ label.
Distributions are similar across data splits.
\tref{tab:AZ_label_counts} also lists counts for \textsc{Abstract}, which we decide to exclude from our modeling experiments because including it resulted in performance decreases due to confusion with other labels.
Locating the abstract in a document can usually be solved in rule-based ways as abstracts of publications are commonly available in a machine-readable format.

During annotation, we introduced two subtypes of \azLabelExplanation, \textsc{Hypothesis} and \textsc{Assumption}, distinguishing between scientific hypotheses and assumptions made by the author in cases where often choices are possible.
As the overall counts and agreement were low, we decided to only use the supertype \azLabelExplanation in all experiments.

\begin{figure*}[ht]
    \centering
    \begin{subfigure}[b]{0.4\textwidth}
    \includegraphics[width=1\textwidth]{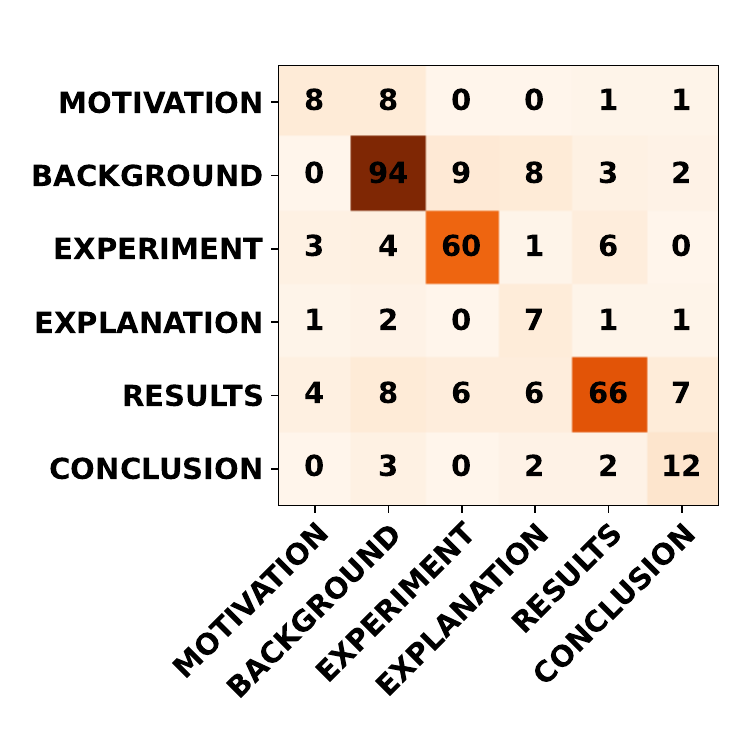}
    \caption{Coincidence matrix for coarse AZ labels.}
    \label{fig:iaa_confmatrix}
    \end{subfigure}
    \begin{subfigure}[b]{0.55\textwidth}
    \includegraphics[width=1\textwidth]{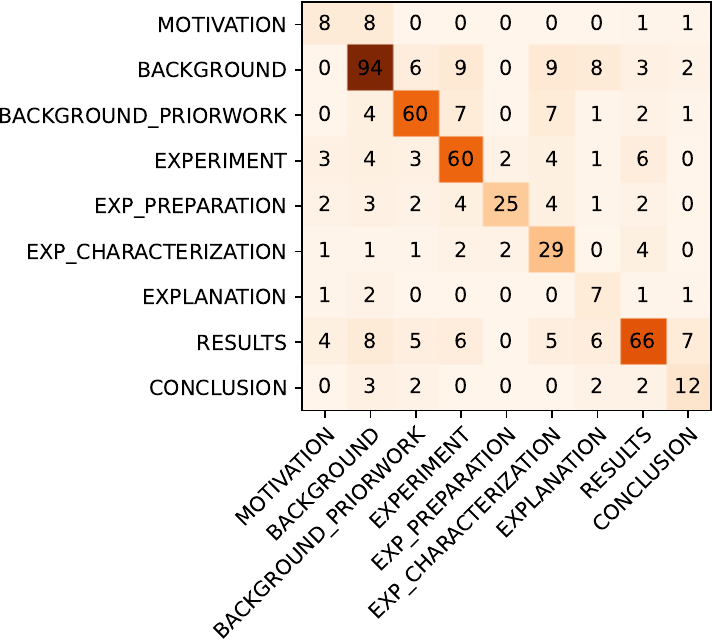}
    \label{fig:iaa_confmatrix2}
    \caption{Coincidence matrix for AZ labels with subtypes.}
    \end{subfigure}
    \caption{Coincidence matrices of inter-annotator agreement study for AZ labels on 357 sentences.}
\end{figure*}

\fref{fig:iaa_confmatrix} shows the label coincidence matrix between the two annotators in the inter-annotator agreement study, i.e., how often each pair of labels co-occurred on an instance.
For all labels except \azLabelMotivation, the majority of coincidences occur on the diagonal.
\azLabelResults is the label most mixed up with others, possibly because these sentences often are long and also contain interpretative information of the other rhetorical types.

\fref{fig:iaa_confmatrix} breaks this information down the level including subtypes.
\azLabelCharacterization and \azLabelPreparation are rarely confused by the domain experts.
Similarly, \azLabelBackground and \azLabelPriorWork are reliably distinguished.

\paragraph{Agreement on sub-labels.}
Our agreement study contained only 12 \azLabelCaption instances.
Data inspection showed that the additional (not the main) annotator neglected to use this tag where appropriate, using only content-related tags on these instances.
There were also not enough instances of the subtypes \azLabelPreparation and \azLabelExperiment{\_\textsc{Characterization}} to measure agreement.
On identifying the subtype \azLabelBackground{\_\textsc{PriorWork}}, annotators achieve a $\kappa$ of 0.8, with (minor) disagreements mainly with regard to using \azLabelBackground or its subtype.

\begin{table}[ht]
\footnotesize
\centering
\begin{tabular}{lrrrr}
\toprule
               \textbf{Label} &  \textbf{total} &  \textbf{train} &  \textbf{dev} &  \textbf{test} \\
\midrule
        \azLabelMotivation &    363 &    273 &   44 &    46 \\
        \azLabelBackground &   3155 &   2423 &  440 &   292 \\
        -\azLabelPriorWork &   1824 &   1387 &  265 &   172 \\
        \azLabelExperiment &   2579 &   1896 &  394 &   289 \\
-\azLabelCharacterization &   1347 &    982 &  200 &   165 \\
-\azLabelPreparation &    962 &    705 &  146 &   111 \\
         \azLabelExplanation &    603 &    430 &   91 &    82 \\
        \azLabelResults &   2953 &   2146 &  440 &   367 \\
    \azLabelConclusion &    680 &    507 &  106 &    67 \\
    \midrule
            \azLabelAbstract &    269 &    190 &   28 &    51 \\
        \azLabelCaption &    485 &    309 &   91 &    85 \\
        \azLabelHeading &    702 &    536 &   96 &    70 \\
        \azLabelMetadata &    210 &    142 &   40 &    28 \\
\bottomrule
\end{tabular}
	\caption{\textbf{Label counts} on the complete dataset and on data split subsets.
		\textbf{Multi-label counts:} Number of sentences in which the label is present. Due to multi-labeling, the sum of these columns exceeds the total amount of sentences. For hierarchical labels, the super-label count includes all sub-label counts.}
	\label{tab:AZ_label_counts}
\end{table}

\paragraph{Agreement on \azLabelHeading.}
As it should be straightforward to identify headings, we looked at the 6 cases that one annotator labeled as \textsc{Heading} but not the other. We found 4 cases to result from broken formatting. One \azLabelMetadata sentence was wrongly labeled \textsc{Heading}, and the remaining \azLabelHeading sentence was missed by the other annotator.

\paragraph{Human \enquote{upper bound}.}
In order to provide a \textit{rough} estimate of how humans would perform on the classification task, we use the data from the agreement study to compute hierarchical precision, recall, and F1 scores.
Due to insufficient data for the remaining labels, we only compute the scores over the following labels: \azLabelHeading, \azLabelMetadata, \azLabelMotivation, \azLabelBackground, \azLabelPriorWork, \azLabelExperiment, \azLabelPreparation, \azLabelCharacterization, \azLabelResults, and \azLabelConclusion.
\tref{tab:human-agreement-upper-bound} reports detailed scores per label.
Scores have been computed using scikit-learn\footnote{\url{https://scikit-learn.org/stable}}.

\begin{table}
\footnotesize
\setlength\tabcolsep{3pt}
\begin{tabular}{lrrrr}
\toprule
{} &  \textbf{Precision} &  \textbf{Recall} &  \textbf{F1} & \textbf{support}\\
\midrule
micro avg.            &       77.3 &    80.0 &      78.7  \\
macro avg.            &       75.0 &    76.1 &      74.9 \\
\midrule
\azLabelHeading              &      100.0 &    81.2 &      89.7 &     32 \\
\azLabelMetadata             &       75.0 &    81.8 &      78.3 &     11 \\
\azLabelMotivation           &       50.0 &    44.4 &      47.1 &     18 \\
\azLabelBackground           &       77.0 &    89.5 &      82.8 &    105 \\
\textsc{-PriorWork} &       77.9 &    90.9 &      83.9 &     66 \\
\azLabelExperiment           &       80.0 &    84.5 &      82.2 &     71 \\
\textsc{-Preparation}     &       92.6 &    73.5 &      82.0 &     34 \\
\textsc{Characterization} &       61.7 &    78.4 &      69.0 &     37 \\
\azLabelResults              &       85.7 &    70.2 &      77.2 &     94 \\
\azLabelConclusion           &       50.0 &    66.7 &      57.1 &     18 \\
\bottomrule
\end{tabular}
\caption{Human agreement computed in terms of hierarchical precision, recall, and F1.}
\label{tab:human-agreement-upper-bound}
\end{table}

\section{Description and Comparison of AZ Datasets.}
\label{sec:datasets-appendix}
In this section, we provide a detailed description and comparison of existing AZ datasets.
The various corpora try to capture very similar information.
However, each corpus defines its set of labels in a slightly different way.
\tref{tab:dataZones} lists the various labels used in the datasets and groups labels used for the same or very similar purpose.
\tref{tab:az_corpora_label_counts} shows the label distributions of the corpora.

\begin{table*}[ht]
	\centering
	\small
	\setlength\tabcolsep{2pt}
	\begin{tabular}{lllllp{5.3cm}}
	\toprule
	\textbf{PubMed} & \textbf{AZ-CL} & \textbf{ART} & \textbf{DRI} & \textbf{\corpusNameShort} &\textbf{Description}\\
	\midrule
	\textsc{Objective} & \textsc{Aim} & \textsc{Hypothesis} & \textsc{Challenge} & \textsc{Motivation} & A sentence describing the research  \\
	& & \textsc{Motivation} & & & target, goal, aim or the motivation\\
	& & \textsc{Goal} & & & for the research.\\
	\midrule
	\textsc{Background} & \textsc{Background}  & \textsc{Background} & \textsc{Background} &  \textsc{Background}&  A statement concerning the knowledge\\
	& \textsc{Contrast} & & & \textsc{PriorWork} & domain or previous related work. \\
	& \textsc{Basis} & \\
	\midrule
	\textsc{Method} & \textsc{Own} & \textsc{Object}, & \textsc{Approach} & \textsc{Experiment} &A sentence describing the research \\
	& & \textsc{Method} & & \textsc{Preparation} &procedure, models used, or observations \\
	&& \textsc{Model} &  &  \textsc{Characteriz.} &made during the research.\\
	&& \textsc{Experiment} & & \textsc{Explanation} \\
	& & \textsc{Observation}\\
	\midrule
	\textsc{Result} & \textsc{Own} & \textsc{Result} & \textsc{Outcome} & \textsc{Results} & A sentence describing the study findings,\\
	&&&&\textsc{Explanation}&effects, consequences, and/or analysis of the results. \\
	\midrule
	\textsc{Conclusion} & \textsc{Own} & \textsc{Conclusion} & \textsc{Outcome} & \textsc{Conclusion} & A statement concerning the support or \\
	& & & \textsc{Futurework} & & rejection of the hypothesis or suggestions of future research. \\
	\midrule
	-- & \textsc{Text} & -- & \textsc{Sentence} & -- & Example sentences, broken sentences, etc. \\
	& \textsc{Other} & & \textsc{Unspecified} & & \\
	\bottomrule
	\end{tabular}
\caption{AZ Corpus Zones Mapping and Descriptions. Compare to \tref{tab:az}.} %
\label{tab:dataZones}
\end{table*}

\begin{table}[ht]
\footnotesize
\centering
\begin{tabular}{llr}
\toprule
\textbf{Dataset} &  \textbf{Label} &  \textbf{Count} \\
                   \midrule
	AZ-CL 	 & \textsc{Own} & 8624 \\
	& \textsc{Other} & 2019 \\
	& \textsc{Background} & 789 \\
	 & \textsc{Contrast} & 600 \\
	& \textsc{Aim} & 313 \\
	 & \textsc{Basis} & 246 \\
	 & \textsc{Text} & 227 \\
\midrule
	ART & \textsc{Result} & 7373 \\
		 & \textsc{Background} & 6657 \\
		& \textsc{Observation} & 4659 \\
	& \textsc{Method} & 3751 \\
		 & \textsc{Model} & 3456 \\
		& \textsc{Conclusion} & 3083 \\
	 & \textsc{Experiment} & 2841 \\
	 	 & \textsc{Object} & 1190 \\
	 	 	 & \textsc{Hypothesis} & 656 \\
	 & \textsc{Goal} & 548 \\
	 & \textsc{Motivation} & 466 \\

    \midrule
	DRI  	 & \textsc{Approach} & 5038 \\
	& \textsc{Background} & 1760 \\ 
	& \textsc{Sentence} & 1247 \\
		 & \textsc{Outcome} & 1175 \\
		& \textsc{Unspecified} & 759 \\
		& \textsc{Challenge} & 351 \\
	 & \textsc{Outcome\_Contribution} & 219 \\ 
	 & \textsc{FutureWork} & 136 \\
	 & \textsc{Challenge\_Hypothesis} & 7 \\
    \midrule
	MatSci PubMed &  \textsc{Results} & 1282 \\
	 & \textsc{Objective} & 1264 \\
	 & \textsc{Methods} & 1198 \\
	 & \textsc{Conclusion} & 380 \\
	& \textsc{Background} & 60 \\
\bottomrule
\end{tabular}
	\caption{Label counts for the different AZ corpora.}
	\label{tab:az_corpora_label_counts}
\end{table}

\paragraph{AZ-CL corpus.}
\label{sec:data:azCorpus}

The Argumentative Zoning \citep[AZ,][]{teufel-etal-1999-annotation, teufel-moens-1999-discourse} corpus\footnote{\url{https://github.com/WING-NUS/RAZ}} consists of 80 manually annotated open-access \textbf{computational linguistics} research articles.
Sentences are marked according to their argumentative zone or rhetorical function as one of the following classes: \textsc{Aim}, \textsc{Background}, \textsc{Basis}, \textsc{Contrast}, \textsc{Other}, \textsc{Own} or \textsc{Text}.
Inter-annotator agreement is reported as substantial ($\kappa$ = 0.71).
The distribution of classes is quite skewed towards \textsc{Other} and \textsc{Own}.

\paragraph{ART corpus.}
\label{sec:data:ARTCorpus}
The ART corpus\footnote{\url{https://www.aber.ac.uk/en/cs/research/cb/projects/art/art-corpus/}} \citep{soldatova2007ontology} covers topics in \textbf{physical chemistry} and \textbf{biochemistry}.
Articles are annotated according to the CISP/CoreSC annotation scheme \cite{liakata2008guidelines}.
Sentences are labeled with one of the categories \textsc{Hypothesis}, \textsc{Motivation}, \textsc{Goal of investigation}, \textsc{Background}, \textsc{Object of investigation}, \textsc{Research method}, \textsc{Model},  \textsc{Experiment}, \textsc{Observation}, \textsc{Result} or \textsc{Conclusion}.
The annotation scheme also defines subcategories for some of these.
The corpus has been annotated by domain experts.
In a preliminary study, $\kappa$ was measured as 0.55, however, for the final corpus, only the annotators that had the highest average agreement were selected.
Hence, the agreement in the final corpus is expected to be higher.

\paragraph{DRI corpus.}
\label{sec:data:DRICorpus}
The Dr. Inventor Multi-Layer Scientific Corpus\footnote{\url{http://sempub.taln.upf.edu/dricorpus}} \citep[DRI,][]{fisas-etal-2016-multi,fisas-etal-2015-discoursive}, contains 40 scientific articles taken from the domain of \textbf{computer graphics}. 
Each of the 10,784 sentences was annotated with one of the rhetorical categories: \textsc{Challenge}, \textsc{Background}, \textsc{Approach}, \textsc{Outcome} or \textsc{Futurework}. They have also included two other categories \textsc{Sentence} for sentences that are characterized by segmentation or character encoding errors and \textsc{Unspecified} for sentences where identification is not possible. 
Also to note was the possibility to annotate a combination of two different categories as seen in the example of: \textsc{Outcome\_Contribution}, \textsc{Challenge\_Goal} and \textsc{Challenge\_Hypothesis}.
Manual annotation reaches a $\kappa$ value of 0.66.

\paragraph{PubMed corpus.}
\label{sec:data:PubmedCorpus}
The PubMed corpus\footnote{\url{https://github.com/dead/rhetorical-structure-pubmed-abstracts}} \citep{deMoura2018using} contains abstracts of papers in the \textbf{biomedical} domain extracted from PUBMED/MEDLINE. The collected abstracts were written in English and annotated with predefined section names by their authors; based on the mapping provided by the U.S. National Library of Medicine (NLM), the section names were collapsed into five rhetorical roles: \textsc{Background}, \textsc{Objective}, \textsc{Methods}, \textsc{Results}, and \textsc{Conclusions}.
The abstracts that did not contain the five mentioned rhetorical roles were removed from the dataset with the resulting corpus containing close to 5 million sentences.
The dataset is not particularly challenging: a simple CRF model achieves an F-score of 93.75, an LSTM-based model achieves 94.77 according to \citet{deMoura2018using}.

\section{Examples}
\label{sec:examples}
In this section, we present and discuss several examples from our dataset.

\subsection{Example Sentences}

\begin{itemize}
    \item \azLabelMotivation: \textit{Therefore, it is highly desirable to develop an innovative technology to raise the mass activity of Ir-based OER catalysts to the targeted level.}
    \item \azLabelBackground: \textit{For photocatalytic water splitting using photoelectrochemical cells (PECs), the charge carriers are created from the photovoltaic effect close to the catalytic site.}
    \item \azLabelPriorWork: \textit{Proton exchange membrane (PEM) electrolysis, which occurs in acidic electrolytes (pH 0–7), has better efficiency and enhanced ramping capability over other types of electrolysis [7].}
    \item \azLabelExperiment: \textit{In order to find an optimum efficiency of the PV–electrolysis, different combinations of the electrolyzer with A-CIGS-based thin film solar cell modules with different band gaps of the cell were examined.}
    \item \azLabelPreparation: \textit{Pre-sputtering was performed for 5 min in argon plasma in order to remove surface impurities.}
    \item \azLabelCharacterization: \textit{The current density-potential (j–V) characteristics of the A-CIGS cells were recorded under simulated AM 1.5G sunlight in a set-up with a halogen lamp (ELH).}
    \item \azLabelExplanation: \textit{A possible explanation for the superior ECSA-specific activity in the 3D WP-structured catalysts is efficient removal of oxygen bubbles from the catalyst layer.}
    \item \azLabelResults: \textit{The load curves were similar for the electrolyzers with different WO3 thin films and the lowest potential needed for 10 mA cm-2 in the overall reaction was 1.77 V.}
    \item \azLabelConclusion: \textit{The Cu-N- rGO demonstrated superior catalytic activity to the counterpart N-rGO, and enhanced durability compared to commercial Pt/C.}
\end{itemize}

Structural tags are used, for example, in the following cases.

\begin{itemize}
    \item \azLabelHeading: \textit{4. Discussion and concluding remarks}
    \item \azLabelMetadata: \textit{This research was funded by Hubei Superior and Distinctive Discipline Group of “Mechatronics and Automobiles” (No.XKQ2019009).}
    \item \azLabelCaption: \textit{Figure 8. Enlarged view of the shaded portion of Figure 7.}
\end{itemize}

\subsection{Multi-Label Examples}
In contrast to earlier works on AZ, our approach to labeling AZ in materials science publications uses a multi-label approach.
In this section, we discuss some multi-label examples.

\begin{itemize}
\item \azLabelBackground, \azLabelPriorWork, \azLabelResults: \textit{This indicates that the HER follows a rate-determining Volmer or Heyrovsky step for different sputtering conditions without any order [40,41].} %
In this example, a result obtained in the current paper confirms a result known from prior work.
\item \azLabelExperiment, \azLabelCharacterization, \azLabelResults, \azLabelExplanation:
\textit{Attributing this enthalpy release exclusively to the removal of grain boundaries in stage B, a specific grain boundary energy(2)γ=Hρ3dini-1-dfin-1=0.85$\pm$0.08Jm-2is estimated using the initial and final crystallite diameters of stage B, as given above (dini=222nm, dfin=764nm), as well as the Cu bulk value of 8.92gcm-3 for the mass density ρ.} %
The first subordinate clause of this sentence (\textit{Attributing ... stage B}) is an \azLabelExplanation. The remainder of the sentence states a \azLabelCharacterization.
\item \azLabelBackground, \azLabelPriorWork, \azLabelResults, \azLabelConclusion: \textit{Furthermore, the fatigue life decreased approximately by  more than 12\% when the pre-corroded time was doubled, and the fatigue life decreased  approximately by more than 11\% when the applied stress level was doubled, indicating that both pre-corroded time and applied stress level can significantly affect the fatigue life of specimens, which shows a good agreement with the previous works [37,38].} This example illustrates a case where our simplification of labeling entire sentences comes to its limits: The first part of the sentence (\textit{Furthermore ... was doubled}) reports \azLabelResults while the second part draws a \azLabelConclusion drawing connections to specific \azLabelPriorWork.
\end{itemize}

\section{Detailed Results}
\label{sec:app-more-results}

In this section, we provide detailed results for the experiments presented in the main part of the paper.

\tref{tab:az_per_label_scores_with_and_wo_oversampling} (no oversampling and ML-ROS) and \tref{tab:az_per_label_scores_multi_task} (multi-task AZ-CL) show the results in terms of precision, recall and (hierarchical) F1 for each label individually.
We report the results on both dev and test of the specific model that performed best on dev compared to all other models.

First, we compare the difference between no oversampling at all and when using ML-ROS.
As shown in \tref{tab:az_labelcounts}, \azLabelMotivation, \azLabelMetadata, and \azLabelCaption are the least frequent labels in our dataset.
Except for \azLabelMetadata on the test set, there is always an increase in terms of F1-score when applying ML-ROS on minority labels during training.
The biggest increase of $5.8$ happened for \azLabelMotivation on the test set.
Furthermore, there is also an improvement of 1.2 points on dev and 2.5 points on test in terms of F1-score for \azLabelExplanation, which is fourth in the list of rarest AZ labels.

During our experimentation, we observed that ML-ROS tends to be especially helpful for models that show strong performance on majority labels, but not on minority labels. 
Other models with different hyperparameters achieve even better scores on minority labels without oversampling; however, they tend to have worse overall performance.

Next, we describe the effects of \textbf{multi-task training} with the AZ-CL dataset.
We also apply ML-ROS to \corpusNameShort in our multi-task experiments.
Both micro-F1 and macro-F1 increase by 0.5 points in terms of micro- and macro-F1 when using multi-tasking instead of ML-ROS only.
Most of the per-label F1-scores increased when using multi-tasking, interestingly with notable differences for \azLabelCharacterization ($4.8$) and \azLabelMetadata ($5.6$).
We conclude that multi-tasking with AZ-CL helps supporting common majority labels even though the domain of this dataset is clearly different from ours.

In contrast, multi-task learning with the other datasets consistently resulted in \textit{decreases} of performance.
The chemistry domain is intuitively closest to that of materials science, hence, we would have expected ART to be a good additional dataset in multi-task learning.
\citet{brack2022cross} provide some insights into cross-domain learning of AZ categories using datasets from biomedicine, chemistry, and computer graphics.
Our \corpusNameShort, alongside AZ-CL, opens up new research opportunities.

\begin{table}[t]
	\footnotesize
	\centering
	\setlength\tabcolsep{2pt}
	\begin{tabular}{llrrr}
		\toprule
		\textbf{Method} & \textbf{LM} & \textbf{mic.-F1} & \textbf{mac.-F1}\\ 
		\midrule
		No Oversampling & BERT & 72.6$_{\pm 1.0}$ & 65.5$_{\pm 0.7}$ \\
		& MatSciBERT & 76.3$_{\pm 0.7}$ & 70.1$_{\pm 0.7}$ \\
		& SciBERT & 76.2$_{\pm 0.9}$ & 70.2$_{\pm 0.6}$ \\
		\midrule
		ML-ROS & SciBERT & 76.7$_{\pm 0.7}$ & 70.6$_{\pm 0.9}$ \\
		+ MT (+PM) %
		& SciBERT & 76.5$_{\pm 0.4}$ & 69.5$_{\pm 0.5}$ \\
		+ MT (+ART) & SciBERT & 75.0$_{\pm 0.9}$ & 68.9$_{\pm 1.1}$\\
		+ MT (+AZ-CL) & SciBERT & \textbf{77.2$_{\pm 0.3}$} & \textbf{71.1$_{\pm 0.5}$} \\
		+ MT (+DRI) & SciBERT & 76.6$_{\pm 0.3}$ & 70.5$_{\pm 0.4}$ \\
		+ MT (+ART+AZ+DRI) & SciBERT & 76.4$_{\pm 0.6}$ & 70.2$_{\pm 0.5}$ \\
		Data Augm. (+PM) %
		&  SciBERT & 77.1$_{\pm 0.8}$ & 70.8$_{\pm 1.3}$ \\
		\midrule
		\textit{human agreement}* & & \textit{78.7}\hspace*{5.5mm} & \textit{74.9}\hspace*{5.5mm}\\
		\bottomrule
	\end{tabular}
	\caption{Results on \corpusNameShort test set, hierarchical micro/macro F1: MT=Multi-Task models, *not directly comparable.}
	\label{tab:az_results}
\end{table}

In addition, we perform a \textbf{data augmentation} experiment using AZ data from scientific abstracts of the PubMed Medline corpus\footnote{\url{https://www.nlm.nih.gov/databases/download/pubmed_medline.html}}, filtering for abstracts that were published in journals related to the materials science domain (see \aref{sec:journal-list}).
We map the four PubMed AZ labels \pubmedLabelBackground, \pubmedLabelObjective, \pubmedLabelResults, and \pubmedLabelConclusions to our four AZ labels \azLabelBackground, \azLabelMotivation, \azLabelResults and \azLabelConclusion.
Augmenting with data from the PubMed Medline dataset also helps to achieve better performance.
However, the micro-F1 score is 0.1 lower and the macro-F1 score is 0.3 lower compared to the MT (+AZ-CL) model.
On the other hand, training is much more time-efficient since a low augmentation percentage of $10\%$ is sufficient to get good results.

\begin{table*}[ht]
\centering
\footnotesize
\begin{tabular}{lrrrrrrrr}
    \toprule
    \multirow{2}{*}{\textbf{Label}} & \multicolumn{3}{c}{\textbf{dev}} && \multicolumn{3}{c}{\textbf{test}}  \\ 
    \cmidrule{2-4} \cmidrule{6-8}
    & \multicolumn{1}{c}{P} & \multicolumn{1}{c}{R} & \multicolumn{1}{c}{H. F1} && \multicolumn{1}{c}{P} & \multicolumn{1}{c}{R} & \multicolumn{1}{c}{H. F1} & \textbf{Count} \\
    \toprule
    \textbf{SciBERT, no oversampling}\\
        \azLabelMotivation & 65.5 &     46.8 &   54.4 & {} & 68.5 &     36.5 &   47.6 & 363 \\
        \azLabelBackground  & 89.2 &     80.0 &   \textbf{84.3} & {} & 85.0 &     76.6 &   80.6 & 3155\\
            -\azLabelPriorWork & 97.0 &     84.5 &   \textbf{90.3} & {} & 92.9 &     67.9 &   78.4 & 1824\\
        \azLabelExperiment & 82.1 &     85.8 &   83.9 & {} & 80.6 &     82.6 &   \textbf{81.6} & 2579\\
            -\azLabelCharacterization  & 72.0 &     68.9 &   \textbf{70.3} & {} & 75.8 &     67.3 &   \textbf{71.1} & 962\\
            -\azLabelPreparation & 65.2 &     65.1 &   65.0 & {} & 78.6 &     69.7 &   \textbf{73.7} & 1347\\
         \azLabelExplanation & 46.3 &     33.0 &   38.4 & {} & 55.0 &     35.9 &   43.4 & 603\\
        \azLabelResults  & 75.0 &     84.6 &   79.5 & {} & 79.9 &     85.9 &   82.8  & 2953 \\
    \azLabelConclusion  & 56.7 &     55.3 &   \textbf{56.0} & {} & 42.4 &     43.0 &   \textbf{42.6} & 680\\
    \midrule
        \azLabelCaption & 92.4 &     75.2 &   82.9 & {} & 80.9 &     68.9 &   74.4 & 485\\
        \azLabelHeading & 84.8 &     97.9 &   90.9& {} & 87.4 &     96.6 &   91.7 & 702\\
        \azLabelMetadata & 93.1 &     68.0 &   78.6 & {} & 78.6 &     72.9 &   \textbf{75.2} & 210\\
        \midrule
        \textit{Average} & \textbf{76.6} & 70.4 & 72.9 & {} & \textbf{75.5} & 67.0 & 70.2\\
    \toprule
    \textbf{SciBERT, ML-ROS}\\
        \azLabelMotivation & 56.3 &     55.9 &   \textbf{55.9} & {} & 72.9 &     43.0 &   \textbf{53.4} & 363\\
        \azLabelBackground  & 82.2 &     84.8 &   83.5 & {} & 79.7 &     84.2 &   \textbf{81.9} & 3155\\
            -\azLabelPriorWork & 96.0 &     84.5 &   89.9 & {} & 90.5 &     71.3 &   \textbf{79.7} & 1824\\
        \azLabelExperiment & 85.1 &     83.2 &   \textbf{84.1} & {} & 81.1 &     81.7 &   81.4 & 2579\\
            -\azLabelCharacterization  & 73.3 &     67.3 &   70.1 & {} & 73.2 &     67.5 &   70.2 & 962\\
            -\azLabelPreparation & 69.4 &     63.4 &   \textbf{66.3} & {} & 73.8 &     69.5 &   71.5 & 1347\\
         \azLabelExplanation & 45.7 &     35.2 &   \textbf{39.6} & {} & 53.4 &     40.2 &  \textbf{ 45.9} & 603\\
        \azLabelResults  & 77.6 &     83.4 &   \textbf{80.4} & {} & 83.6 &     83.8 &   \textbf{83.7} & 2953\\
    \azLabelConclusion  & 60.6 &     44.5 &   51.3 & {} & 46.8 &     35.2 &   40.1 & 680\\
    \midrule
        \azLabelCaption & 91.7 &     79.6 &   \textbf{85.2} & {} & 77.9 &     73.6 &   \textbf{75.7} & 485\\
        \azLabelHeading & 85.4 &     97.5 &   \textbf{91.1} & {} & 90.6 &     96.3 &   \textbf{93.4} & 702\\
        \azLabelMetadata & 89.3 &     70.5 &   \textbf{78.8} & {} & 61.9 &     80.0 &   69.8 & 210\\
        \midrule
        \textit{Average} & 76.1 &     \textbf{70.8} &   \textbf{73.0}  & {} & 73.8 &     \textbf{68.9}  & \textbf{70.6} & {} \\
    \bottomrule
\end{tabular}
\caption{Per label scores on dev and test of \corpusNameShort in terms of precision, recall, and hierarchical F1. \textbf{Bold}: best result for label. P, R, and F1 scores are averages over the P, R, F1 scores of 5 folds each.}
\label{tab:az_per_label_scores_with_and_wo_oversampling}
\end{table*}

\begin{table*}[ht]
\centering
\footnotesize
\begin{tabular}{lccccccc}
    \toprule
    \multirow{2}{*}{\textbf{Label}} & \multicolumn{3}{c}{\textbf{dev}} && \multicolumn{3}{c}{\textbf{test}}  \\ 
    \cmidrule{2-4} \cmidrule{6-8}
    & \multicolumn{1}{c}{P} & \multicolumn{1}{c}{R} & \multicolumn{1}{c}{H. F1} && \multicolumn{1}{c}{P} & \multicolumn{1}{c}{R} & \multicolumn{1}{c}{H. F1}  \\ 
    \midrule
        \azLabelMotivation & 62.7 &     54.1 &   58.0 & {} & 71.2 &     43.9 &   54.3 \\
        \azLabelBackground  & 85.6 &     82.1 &   83.8 & {} & 80.9 &     81.6 &   81.2 \\
            -\azLabelPriorWork & 95.4 &     84.2 &   89.4 & {} & 93.7 &     68.8 &   79.3 \\
        \azLabelExperiment & 83.6 &     82.8 &   83.2 & {} & 83.1 &     83.0 &   83.0 \\
            -\azLabelCharacterization  & 73.7 &     65.9 &   69.3 & {} & 77.4 &     73.0 &   75.0 \\
            -\azLabelPreparation & 69.4 &     55.6 &   61.7 & {} & 79.4 &     67.2 &   72.8 \\
         \azLabelExplanation & 42.6 &     35.8 &   38.8 & {} & 51.2 &     35.9 &   41.7 \\
        \azLabelResults  & 76.6 &     84.4 &   80.3 & {} & 81.5 &     85.1 &   83.2 \\
    \azLabelConclusion  & 61.8 &     49.6 &   55.0 & {} & 41.0 &     32.8 &   36.4 \\
    \midrule
        \azLabelCaption & 90.5 &     77.6 &   83.5 & {} & 79.2 &     76.2 &   77.7 \\
        \azLabelHeading & 84.7 &     97.7 &   90.7 & {} & 88.9 &     97.4 &   92.9 \\
        \azLabelMetadata & 84.3 &     72.0 &   77.6 & {} & 70.6 &     81.4 &   75.4 \\
        \midrule
    \bottomrule
\end{tabular}
\captionsetup{justification=centering}
\caption{Per label scores on dev and test in terms of precision, recall, and hierarchical F1 using multi-task learning with the AZ-CL dataset, SciBERT, ML-ROS.}
\label{tab:az_per_label_scores_multi_task}
\end{table*}

\end{document}